\documentclass{article}

\usepackage{PRIMEarxiv}

\usepackage[utf8]{inputenc} 
\usepackage[T1]{fontenc}    
\usepackage{hyperref}       
\usepackage{url}            
\usepackage{booktabs}       
\usepackage{amsfonts}       
\usepackage{nicefrac}       
\usepackage{microtype}      
\usepackage{lipsum}
\usepackage{amsmath}
\usepackage{appendix}
\usepackage{longtable}
\usepackage{multirow, booktabs, bbding, wrapfig}
\usepackage[table]{xcolor}
\usepackage{fancyhdr}       
\usepackage{graphicx}       
\graphicspath{{media/}}     

\hypersetup{hidelinks}
\title{Furthest Reasoning with Plan Assessment: Stable Reasoning Path with Retrieval-Augmented Large Language Models
}

\author{
  Yin Zhu \footnotemark[1] \\
  Nanjing University \\
  \texttt{yinzhu@smail.nju.edu.cn} \\
   \And
  Zhiling Luo \footnotemark[2] \\
  DAMO Academy, Alibaba Group \\
  \texttt{godot.lzl@alibaba-inc.com} \\
  \And
  Gong Cheng\\
  Nanjing University\\
  \texttt{gcheng@nju.edu.cn}
}

\begin{document}
\maketitle
\footnotetext[1]{Work done during an internship at DAMO Academy, Alibaba Group}
\footnotetext[2]{Corresponding author}

\begin{abstract}
Large Language Models (LLMs), acting as a powerful reasoner and generator, exhibit extraordinary performance across various natural language tasks, such as question answering (QA).
Among these tasks, Multi-Hop Question Answering (MHQA) stands as a widely discussed category, necessitating seamless integration between LLMs and the retrieval of external knowledge. 
Existing methods employ LLM to generate reasoning paths and plans, and utilize IR to iteratively retrieve related knowledge, but these approaches have inherent flaws.
On one hand, Information Retriever (IR) is hindered by the low quality of generated queries by LLM. 
On the other hand, LLM is easily misguided by the irrelevant knowledge by IR.
These inaccuracies, accumulated by the iterative interaction between IR and LLM, lead to a disaster in effectiveness at the end.
To overcome above barriers, in this paper, we propose a novel pipeline for MHQA called Furthest-Reasoning-with-Plan-Assessment (FuRePA), including an improved framework (Furthest Reasoning) and an attached module (Plan Assessor).
1) Furthest reasoning operates by masking previous reasoning path and generated queries for LLM, encouraging LLM generating chain of thought from scratch in each iteration.
This approach enables LLM to break the shackle built by previous misleading thoughts and queries (if any).
2) The Plan Assessor is a trained evaluator that selects an appropriate plan from a group of candidate plans proposed by LLM.
Our methods are evaluated on three highly recognized public multi-hop question answering datasets and outperform state-of-the-art on most metrics (achieving a 10\%-12\% in answer accuracy).
\end{abstract}


\section{Introduction} \label{sec:introduction}
The problem of Question Answering (QA) has long attracted considerable research attention. As progressively improved models provide excellent performance on simple QA tasks, researchers have recently turned their attention to more complex questions. Among these, Multi-Hop QA (MHQA) has shown enormous potential and challenge over the years. (\cite{yang2018hotpotqa}, \cite{chen2020hybridqa}, \cite{welbl2018constructing}, \cite{kovcisky2018narrativeqa}, \cite{mihaylov2018can}, \cite{mavi2022survey} etc.) To solve complicated multi-hop questions, the ability of multi-step reasoning and retrieval is essential and required, thus promoting the comprehensive improvement of natural language processing models. In this complex task, a continuous and accessible reasoning path is crucial for accuracy and interpretability, as the answer is obtained after combining several reasoning steps.

With the rapid rise of powerful Large Language Models (LLMs) and their CoT \cite{wei2022chain} capability, an intuitive and widely implemented basic method for solving MHQA is to use LLMs and IRs alternately. Typically, at each step, LLM takes previous reasoning steps, executes queries and retrieves evidence, and then outputs the next query. IR then uses this query to retrieve the relevant information. This loop continues until LLM has obtained all the necessary evidence and answered the question. \cite{yao2022react}, \cite{xu2023search}, \cite{shao2023enhancing}, \cite{chen2023chatcot} apply this basic framework. Despite the amazing performance of LLM on natural language tasks, there are some difficulties in solving MHQA with retriever-augmented methods.

\begin{figure}
    \centering
    \includegraphics[width=1.0\textwidth]{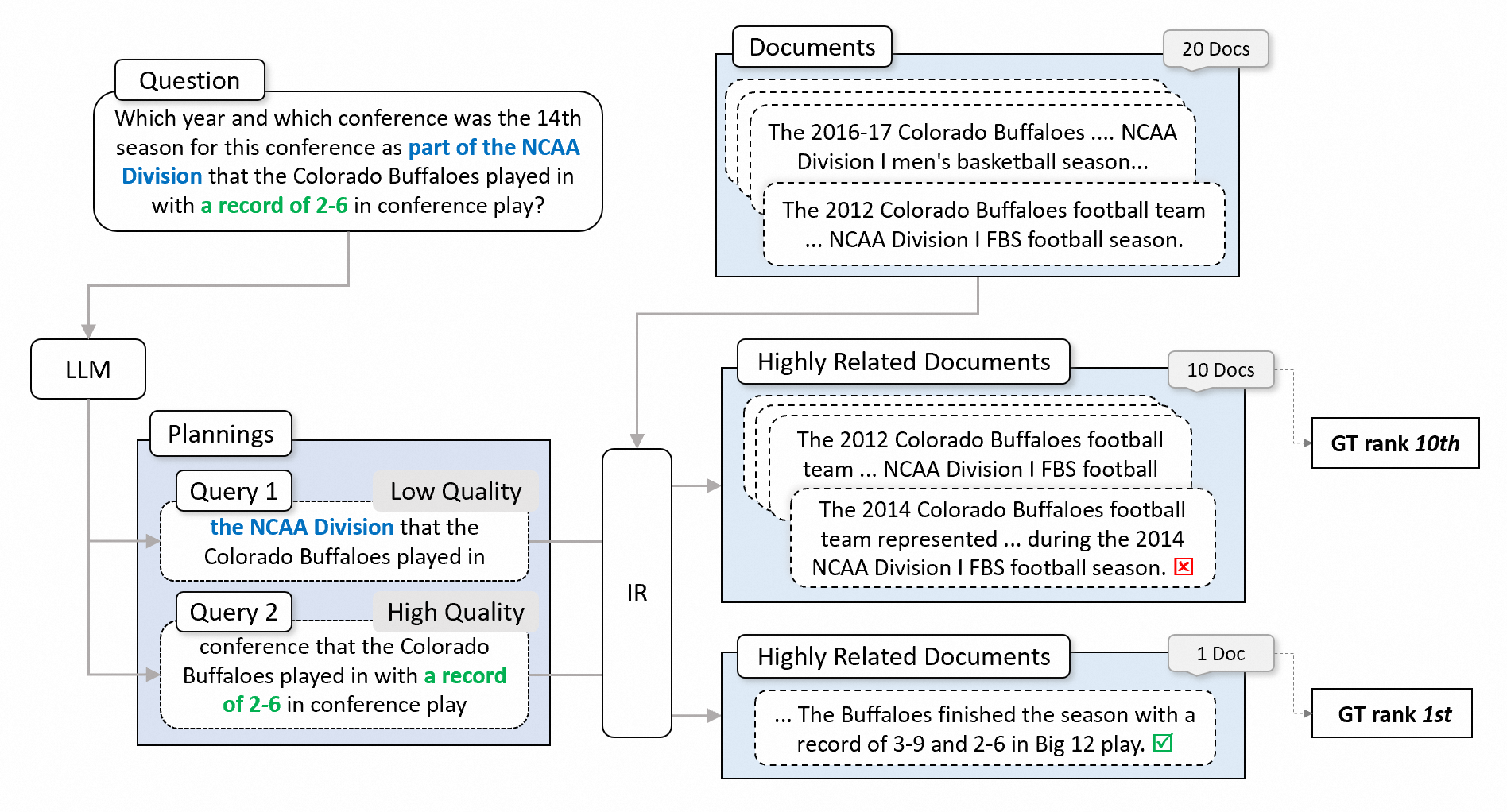}
    \caption{Instance that shows importance of selecting high-quality query. Information Retriever detects relevant documents from corpus with the query. First query recalls the ground truth at $10th$ because it's not distinctive and several highly related documents confuses IR. Second query of high quality recalls ground truth at top-$1$ with less highly related documents due to its distinction among the corpus.}
    \label{fig:instance}
\end{figure}

The first major challenge lies in the evaluation and selection of \textbf{high-quality query} decomposed by LLM. More specifically, with the ability of generation diversity, LLM makes a group of available candidate plannings. Different plannings raise the ceiling of high-quality queries, but low-quality queries also appear in the group, which need to be eliminated. If IR takes the low-quality query, the evidence retrieved is very likely to be irrelevant or unhelpful. In the next iteration, LLM trusts this false evidence and makes even worse plans that deviate significantly from the correct reasoning path. For example, as shown in Figure~\ref{fig:instance}, LLM generates some different queries for the original question. The first query focuses on \emph{``the NCAA Division''}, while the second query emphasises information about \emph{``a record of 2-6''}. During the retrieval process, it turns out that huge documents describe Colorado Buffaloes teams of different sports and seasons in the NCAA division, so the first query lacks distinction and leads to wrong retrieved evidence. The corresponding ground truth ranks $10th$, lower than many other documents containing \emph{``the NCAA Division''}. On the contrary, the second query has the distinctive content, which makes it easy for IR to select the target document. However, existing methods do not take advantage of generation diversity and rely heavily on the single generated query of LLM without any evaluation, which causes irreversible damage. As an additive mechanism, self-consistency \cite{wang2022self} is mainly targeted at tasks where a diverse set of reasoning paths could directly lead to the same results without any query in the middle of the reasoning steps.

Repetitive interaction between LLM and IR interrupts the CoT reasoning path, which is important in MHQA, so \textbf{maintain reasoning path continuity} is another important challenge. The expected reasoning path deviates at two stages, LLM planning and IR retrieval, because LLM and IR are both black boxes in the solution process. First, when taking previous reasoning content, queries and evidence as input, LLM makes incorrect planning due to hallucination and forgetting. Second, although the query is good enough, IR still fails to acquire the correct evidence. As conversation and reasoning progress, planning and retrieval deviations occur from time to time. Researchers have made considerable efforts to address this challenge with various ways to avoid breaking the chain of reasoning. ChatCot \cite{chen2023chatcot} incorporates LLM and IR with multi-turn conversations to prevent tool interruptions from affecting the continuity of reasoning in the chain of thought. It is worth noting that ChatCot assesses the value of retrieved documents to help clarify the reasoning process for LLM. However, in a few-shot setting, too many rounds of conversation inevitably lead to neglect of previous reasoning or instructions, thus causing a divergent reasoning path. In \cite{xu2023search}, LLM generates a global reasoning chain at each iteration to avoid discontinuity, but suffers from the same problem as ChatCot. Iter-RetGen \cite{shao2023enhancing} takes each iteration as a complete QA process. This method constructs an intact reasoning path at each iteration with little risk of ignorance, but without any adaptive mechanism such as an early stop or confidence threshold, excessive iterations and supporting documents can lead to uncertain reasoning paths and worse results, as shown.

\begin{figure}
    \centering
    \includegraphics[width=1.0\textwidth]{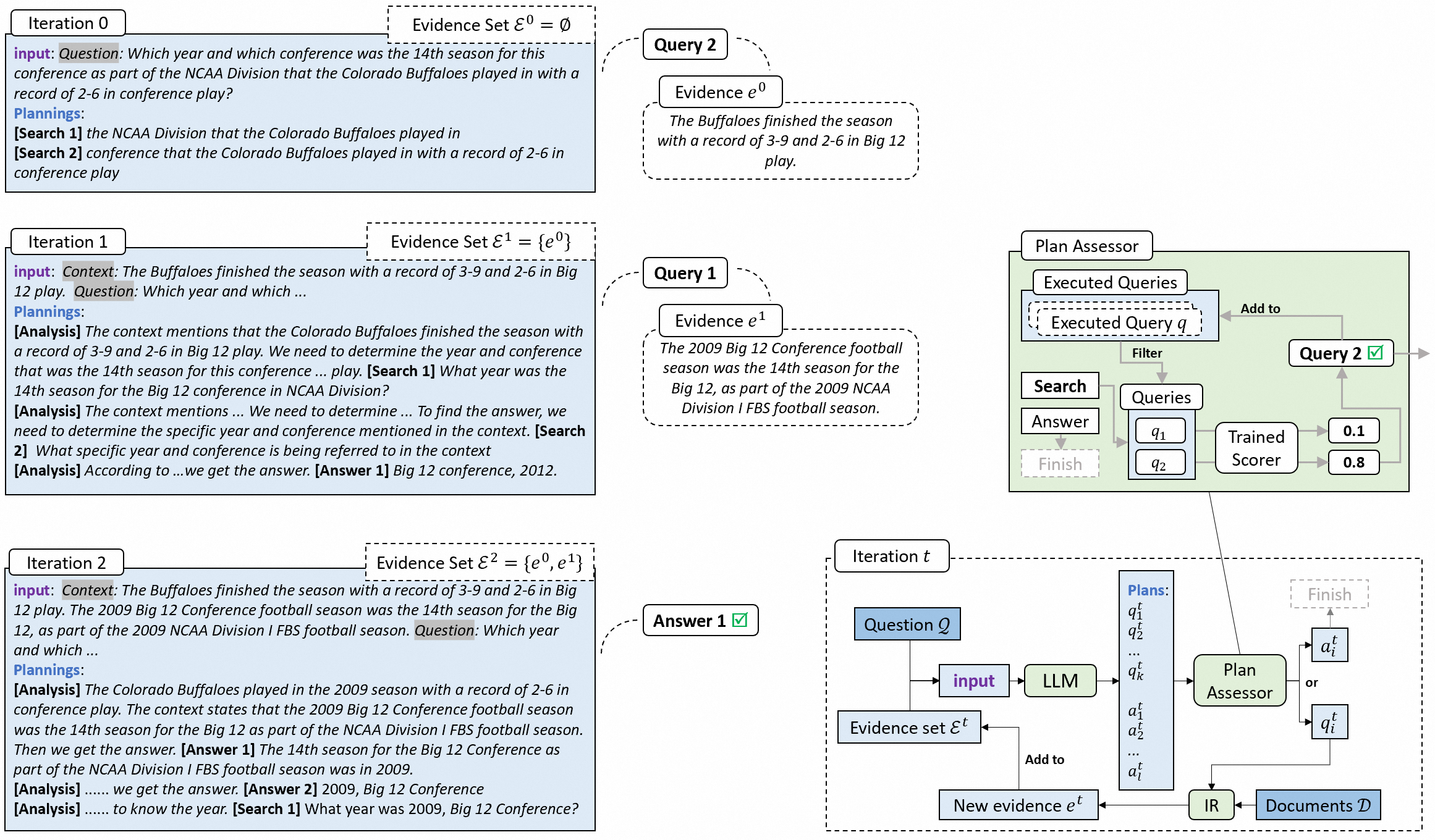}
    \caption{Iteration details in our framework are shown in the right part. In each iteration $t$, LLM generates reasoning chain and plans (e.g. query or answer) with retrieved context and original question as input. Module plan assessor handles the plannings and chooses the best one. Left part is the solving process of example of Figure~\ref{fig:instance}}
    \label{fig:pipeline}
\end{figure}

In this paper, we propose a framework called Furthest Reasoning with Plan Assessment (FuRePA) to address the above challenges (Figure~\ref{fig:pipeline}). 
At each iteration $t$, LLM has access only to the retrieved evidence and the original question, and then predicts the next plans through in-context learning. Unlike conventional interactive methods, the previous chains of reasoning and questions are always masked, forcing LLM to reason from the beginning to the end. Through this mechanism, the LLM analyses the context and reasons without the influence of the previous path. The reasoning process also recovers from false evidence caused by IR.

Plannings include answer determination when all required documents are presented, and query execution when more relevant information is needed. A trained plan assessor (right part of Figure~\ref{fig:pipeline}) then analyses the candidate plans and selects the best one. For different queries, the scorer in the assessor estimates their MRR with respect to the sampled closed-domain context. The query with the highest score is then executed to retrieve a new document.
As the example in the figure shows, at iteration $0$ some queries are generated and the plan assessor selects the second one. The retrieved evidence $e^0$ is added to the set. Then, at iteration $1$, LLM is fed only evidence and the original question and makes some plans. The first question \emph{``what year was the 14th season ...''} is used to retrieve evidence $e^1$. At iteration $2$ the plan evaluator chooses to answer the question and finishes the process.

We evaluate our framework on three multi-hop tasks. Our method gains about 10\% on most  metrics over the state of the art in a few-shot setting. With the ablation study and the case study, our pipeline and optimisation strategy draw conclusions as expected. The masking mechanism and scratch reasoning in our framework effectively handle the aforementioned challenge of reasoning path continuity, as well as avoid forgetting the LLM and improve the reasoning process. The plan evaluation procedure generates and selects plans in a more creative way, including better rewriting and more accurate queries.

We summarize the contributions as follows.
\begin{itemize}
\item A trained plan assessor is a module that estimates the MRR score and selects the \textbf{high-quality query}, which recalls similar texts. High quality searches improve retrieval and identify documents containing potential evidence.
\item A framework named Furthest Reasoning \textbf{maintains reasoning path continuity} and effectively mitigates harm from previous chain of thought or query to the minimum by masking them.
\end{itemize}

\section{Related Work}
\paragraph{Chain of Thought Prompting}
Since large language models offer exciting prospect of in-context few-shot learning through various promptings, Chain of Thought \cite{wei2022chain} is proposed to improve reasoning ability by clarifying intermediate reasoning steps. Following the pioneering work, more works on eliciting reasoning in LLM have been researched. "Step by Step" \cite{kojima2022large} prompting is proved to be effective under zero-shot setting. Auto-CoT \cite{zhang2022automatic} automates the construction of presentations, builds demonstrations by sampling rich questions and generates corresponding reasoning chains instead of manual design. Self-consistency \cite{wang2022self} is a decoding strategy that uses ensemble to replace greedy search on the basis of few-shot-cot to improve decoding accuracy. Least-to-Most \cite{zhou2022least} guides the model to decompose the problem into sub-questions and answer the sub-questions iteratively at each step. Similar methods include Self-Ask \cite{press2022measuring}, ReAct \cite{yao2022react} etc, which all apply to multi-hop questions. To conclude, chain of thought, as a method to introduce structural and logical coherence in generative models, expands the application of LLM to various tasks. By introducing targeted hints, it is expected to improve the generating quality and play an important role in many complex natural language processing tasks, most of which require a series of reasoning steps.

\paragraph{Query Decomposition}
Query decomposition and optimization is crucial in MHQA and other multi-step reasoning tasks. \cite{khot2022decomposed} decomposes tough problems into simpler sub-tasks. LLMs in the decomposer and sub-task handlers take corresponding few-shot prompts. Plan-and-Solve \cite{wang2023plan} also makes decomposition at the beginning and carries it out.
HPE \cite{liu2023answering} parses the multi-hop question into a structured intermediate representation composed of simple questions, representing the relationships. \cite{radhakrishnan2023question} attempts to explore more effective methods than CoT for eliciting faithful reasoning through factored decomposition and re-composition. Nevertheless,
most of the methods above only make LLMs generate the queries and trust them without any optimization or evaluations. Actually, LLMs sometimes make unsatisfying plans, which are not beneficial to following steps.

\paragraph{Retrieval-augmentation}
The idea of retrieval-augmentaion is widely used in all kinds of models. For example, Realm \cite{guu2020retrieval} explored open-domain extractive question answering, and models like RAG \cite{lewis2020retrieval} bring retrieval into generative models. Thanks to rapidly developing LLMs, different methods of retrieval-augmented large language models provide promising solutions for complex knowledge-intensive QA tasks. In-context RALM \cite{ram2023context} and REPLUG \cite{shi2023replug} leave the LM architecture unchanged and prepend retrieved documents to the input. 

Only retrieval of related documents has positive effect, while those irrelevant documents can also harm the quality of generation and reasoning. \cite{lyu2023improving} evaluates data importance and quality of retrieved data points. \cite{ma2023query} trains the query rewriter in retrieval with policy gradient reinforcement learning. In addition, the newly added retrieval module in the pipeline brings resource computing pressure (not only the retrieval requires resources, but the generated model also consumes more resources due to the need to consider the retrieval results), which also requires further evaluated.

When introducing external information into language models, recent works didn't try to mitigate the impact of irrelevant documents when they are retrieved and trusted. To leverage the improvement on retrieval quality and large computing cost in LLMs, our work makes effort to undermine the inevitable effect from IR, avoiding more error accumulation in the following procedure.

\section{Method}
The original question $\mathcal{Q}$ and a retrieval corpus $\mathcal{D} = \{d\}$ serve as the input of the MHQA instance. 

\subsection{Furthest Reasoning} \label{furthest reasoning}
In each iteration $t$ with upper limit $T$, LLM $\mathcal{M}$ accepts retrieved evidences $\mathcal{E}^t$ and the original question $\mathcal{Q}$ integrated as input. With regard of the temperature $tp$ and $n$ choices in the parameters, LLM produces an output $\mathcal{P}^t = \{p_i\mid 1 \leq i \leq n\}$, including $n$ candidate plannings. The circulation is shown in Figure~\ref{fig:pipeline}. The input and output of LLM are represented as:
\begin{equation}
    \mathcal{P}^t = \mathcal{M}(\mathrm{prompt}(\mathcal{E}^t, \mathcal{Q}, \mathrm{sgn}(t)), tp_0+\Delta{tp}, n), \forall 0 \leq t < T
\end{equation}

$\mathrm{sgn}(t)$ is a sign function, which tells whether the step is the first. The prompts at the beginning differ from the latter ones because we definitely cannot get the answer without any retrieved documents. Therefore, the prompts in the first iteration are designed to take original question and output queries only. 

\subsubsection{Prompting}

Since the round of reasoning $\mathrm{sgn}(t)$ is known, the detailed prompting strategy is implemented on the retrieved evidences $\mathcal{E}^t$ and question $\mathcal{Q}$. For example, the system instruction template at latter steps is:

\textbf{\textit{System:}} \emph{``You are given problems requiring multi-step searching and reasoning with related contexts. You should tell me the answer if all necessary information is provided in the context, otherwise tell me what specific information to search next. Don't answer with uncertainty. Lines only start with '[Analysis]' or '[Answer]' or '[Search]'.''} 

This instruction requires LLM to understand the context (retrieved evidences here) and question, and produce the continuous reasoning chain from start to the furthest possible state, followed by planning like answer or query. The planning is generated through chain of thought, consisting of a reasoning chain $r$ and the following plan $q$ or $a$. The exemplars and the expected response patterns are as follows:

\textbf{\textit{User:}} \emph{``Context: $\mathcal{E}^t$ Question: $\mathcal{Q}$''}

\textbf{\textit{Assistant:}} \emph{``[Analysis] $r$ [Search] $q$''} or \emph{``[Analysis] $r$ [Answer] $a$''}

Here we prompt the LLM to plan for the next step only with retrieved evidences and the original question. The output of LLM is a group of plannings $\mathcal{P}^t$ defined by:

\begin{equation}
    \begin{aligned}
        \mathcal{P}^t &:= \{p^t_1, p^t_2, \cdots, p^t_n\} \\
        &= \{(r^t_1, q^t_1), (r^t_2, q^t_2), \cdots, (r^t_k, q^t_k)\} \cup \{(r^t_{k+1}, a^t_1), (r^t_{k+2}, a^t_2), \cdots, (r^t_{n}, a^t_l)\} \\
        &= Q^t \cup A^t
    \end{aligned}
\end{equation}
Unlike many other general methods, we mask the previous reasoning content and executed query because former reasoning and plans have negative effect on the multi-step reasoning process, which is proved in section \ref{sec: experiments}. Thus, LLM will analyze the multi-hop question from scratch according to retrieved evidences, no matter whether they help or not, preventing new plans affected by erroneous previous reasoning and evidences. It is noteworthy that in the few-shot examples, we teach LLM to distinguish similar but irrelevant evidences from truly demanded ones, which further mitigates the redundancy information caused by IR.
The output plannings above include $k$ query executions $\{q^t_i \mid 1 \leq i \leq k \}$ and $l$ answer determinations $\{a^t_j \mid 1 \leq j \leq l \}$ in iteration $t$. Each planning is generated through CoT $r^t_i$.

\subsubsection{Iteration for Furthest Reasoning}
We iterate the furthest reasoning step for at most $T$ times. After we get $\mathcal{P}^t$ at step $t$, the plan assessor (section \ref{plan assessor}) is used to select the best planning $p_{best}$ among them. If $p_{best}$ claims the final answer $a$, then the whole iteration ends with reasoning path $r$ and answer $a$. Otherwise $p_{best}$ is a query execution $q$. Retriever module sorts the documents in $\mathcal{D}$ and adds the top new document $d$ into retrieved evidence set $\mathcal{E}^t$, which is defined by: 
\begin{equation}
    \mathcal{E}^t = 
\begin{cases}
    \emptyset & \text{if } t=0\\
    \mathcal{E}^{t-1} \cup \{d\}, \text{where } d \not\in \mathcal{E}^{t-1} \text{ and } d \in \mathcal{D}, & \text{otherwise.}
\end{cases}
\end{equation}
Then the next iteration runs with new retrieved context $\mathcal{E}^{t+1}$. The answer will be forcibly generated if not determined after last round.

\subsection{Plan Assessor} \label{plan assessor}

The plan assessor $\mathcal{A}$ is a module that takes the various candidate plannings $\mathcal{P}^t$ generated by LLM and outputs the best planning $p_{best}$:
\begin{equation}
    p_{best}=\mathcal{A}(t, \mathcal{P}^t, E^t, \mathcal{D})
\end{equation}
In each iteration in Figure~\ref{fig:pipeline}, LLM gives various chains of thought and subsequent plans, including query execution and answer determination. What plan assessor do is to elect the most suitable one, so that the probability of low quality query and randomly speculated answer is reduced and the whole reasoning process becomes more convincible. $E^t$ is the set of executed queries.

\subsubsection{Planning Decision}
Apart from the first step which is impossible to induce the answer, the candidate plannings may include answer determinations and query executions as shown in section \ref{furthest reasoning}. The planning decision is made by voting algorithm. When the proportion of the answer plannings in $\mathcal{P}^t$ surpasses the certain threshold $\theta$, the reasoning process ends with one pair of complete reasoning path and answer. If not, the rest plannings of query execution are sifted and ranked by query filter (section \ref{query filter}) and scorer (section \ref{query scorer}). At iteration $2$ in Figure~\ref{fig:pipeline}, when most of the plans provide the final answer as \emph{``2009, Big 12 Conference''}, any of the answer plan and its CoT is taken as the outcome. The function is expressed as:
\begin{equation}
    \mathrm{PlanDecision}(t, \mathcal{P}^t) = 
\begin{cases}
    A^t_0, & \text{if } t>0 \text{ and } \frac{\left| A^t \right|}{\left| \mathcal{P}^t \right|} \geq \theta \\
    Q^t, & \text{otherwise}
\end{cases}
\end{equation}

\subsubsection{Query Filter} \label{query filter}

We filter the query plannings $Q^t$ and get $\hat{Q^t}$ with density-based clustering method to eliminate the same queries as before. $\hat{Q^t}$ consists of the centered examples from each cluster. In cases $\hat{Q^t}$ is empty, which means similar queries were executed before, this iteration ends with the temperature increasing $\Delta{tp}$. When identical queries are generated repetitively, which means procedure suffers from low-quality queries or absence of evidences. Temperature raise (decoding probability threshold change) replaces the greedy decoding used in chain-of-thought prompting with more diverse reasoning paths and plans. With this duplication detection and reasoning diversity, LLM can generate alternated contents like rewriting or entity replacement.

\subsubsection{Query Scorer} \label{query scorer}
The query scorer calculates the score representing the quality of a query in a certain context:
\begin{equation}
    s = \mathrm{QueryScorer}(q, \mathcal{D})
\end{equation}

\paragraph{Sampling and Labeling}
We sampled all training data the same way as furthest reasoning in section \ref{furthest reasoning}. For each query generated at different steps, IR scores and ranks the similarity between query and documents directly. The dataset does not provide the exact match document of sub-query, so the ground truth is taken as the first gold document $d_g$ in sorted documents. We label documents from head to the ground truth as positive ones (or highly related ones in Figure~\ref{fig:instance}) ($\hat{y_j}, 1 \leq j \leq g$) , and documents from ground truth to the end as negative ($\hat{y_j}, g < j \leq \left| \mathcal{D} \right|$) because we will use Mean Reciprocal Rank (MRR)\cite{voorhees1999trec} for reference in scoring system. The gold score of a query $\hat{s}$ is the reciprocal of the ground truth location $1/g$.

In the example shown in section \ref{sec:introduction}, if the first query \emph{``the NCAA Division''} is executed, the top documents after IR ranking will be \emph{``The 2014 Colorado Buffaloes ...''} and \emph{``The 2012 Colorado Buffaloes football ...''}, while the most relevant ground truth \emph{``The 2009 Colorado Buffaloes football team represented the University of Colorado in the 2009 NCAA Division I FBS college football season.''} is ranked $10th$. In this case, the quality score of this query is $1/10$ and the documents ranked before ground truth are all labeled as true. Our neural network will be trained to fit the labels. When the model predicts the quality of a query, the estimated true documents will be detected as highly related and confusing ones and undermine the score of that query. We guarantee the query as much as accurate and distinctive with this estimation.

\paragraph{Scoring} \label{scoring}
We train and optimize the scorer with the following process.
\begin{gather}
        \mathbf{H}^{j}=\mathtt{Encoder}(q; d_j), s_j=\sigma(\mathbf{H}^{j}_{\text{[CLS]}}\mathbf{w}_s) \\
        s^{train}=\frac{1}{\sum _{j=1}^{N} s_j} \label{training score}
\end{gather}
The input query $q$ is encoded with each document $d_j$ and projected to a scalar $s_j$, where $\mathbf{H}^j$ is the last hidden states for ``$\text{[CLS]}q\text{[SEP]}d_j$'' and $\mathbf{H}^{j}_{\text{[CLS]}}$ is the hidden states of [CLS] token. The reciprocal of the sum of each scalar $s_j$ is the MRR score of query $q$ in training process. The loss is expressed as follows. BCELoss is used for the document labels and MSELoss is for regression fitting of MRR score. The introduction of MSELoss is to help supervise and predict the document labels better.
\begin{equation}
    \mathcal{L}=\mathrm{BCELoss}(s_j, \hat{y_j}) + \alpha \cdot \mathrm{MSELoss}(s^{train}, \hat s)
\end{equation}
In predicting phase, the score are calculated differently from Formula \ref{training score}. Only the documents whose score exceeds $0.5$ will be predicted as true. In addition, query score will be $0$ if all documents are negative because sometimes when an invalid query is executed, the sum of $s_j$ will be about the same as zero, which enlarges its reciprocal score $s_{pred}$. For example, in iteration $1$ in Figure~\ref{fig:pipeline}, the second query \emph{``What specific year and conference is being referred to in the context''} is an abstract description pointing at no concrete information. In this case $s_j$ of documents in $\mathcal{D}$ are nearly zero, $s^{train}$ will be a huge number which poorly represents the real quality of the query. 
\begin{equation}
    s^{pred}= 
\begin{cases}
    0 &  \sum_{j=1}^{N} \mathrm{sgn}(2 \cdot s_j-1) = 0 \\
    \frac{1}{\sum_{j=1}^{N} \mathrm{sgn}(2 \cdot s_j-1)} &  \sum_{j=1}^{N} \mathrm{sgn}(2 \cdot s_j-1) > 0
\end{cases}
\end{equation}
At last, queries in $\hat{Q^t}$ are sorted by their scores. The highest one is reckoned as the best query $q_{best}$ and added to $E^t$.

\section{Experiments} \label{sec: experiments}

\subsection{Experimental Settings}
\paragraph{Datasets and Evaluation Metrics} We experimented on three recognized MHQA datasets that require information retrieving process: HotPotQA\cite{yang2018hotpotqa}, 2WikiMultiHopQA\cite{ho2020constructing}, and MuSiQue\cite{trivedi2022musique}. The example analyzed in section \ref{sec:introduction} is from HotPotQA. As for the evaluation metrics, we adopt the offical metrics and code in each dataset. Besides, $\mathrm{Acc}^{\dag}$ from Iter-RetGen\cite{shao2023enhancing} and cover-EM\cite{rosset2020knowledge} are also added for comparison.
\paragraph{Baselines}
Our baselines are some recent methods that introduces IR to LLM and explores their interaction. Apart from the basic LLM reasoning baseline (Direct Retrieval and CoT), others are designed for tasks involving multi-step interaction between LLM and IR (ChatCot\cite{chen2023chatcot}, Iter-RetGen\cite{shao2023enhancing}, SearChain\cite{xu2023search}). All methods in our implementation use the same retriever. We didn't experiment with ReAct\cite{yao2022react}, Self-Ask\cite{press2022measuring} and similar methods for two reasons. First, results of these universal prompting methods are shown in papers like Iter-RetGen. Besides, they are conceptually similar to ChatCot and No mask setting (section \ref{sec: ablation study}) on our tasks, which are experimented and compared. In addition, Iter-RetGen does not provide related code and prompts, therefore its results with training are directly taken from its paper, while others are reproduced under our settings.
\paragraph{Implementation Details}
The large language model used in our experiment, including $\mathrm{Acc}^{\dag}$ in evaluation metric is \textit{gpt-3.5-turbo} provided by OpenAI. The information retriever module is TART \cite{asai2022task} (following Iter-RetGen). For all three datasets, the experiments are done on \textit{top 1/10} of dev set under distractor setting. (\textit{1/10} due to the large cost on API call, and proved to be distributed similarly with whole dev set)

In furthest reasoning framework, the initial temperature $tp_0$ of LLM is \textit{0.2/1.0} and the increase $\Delta{tp}$ is \textit{0.8/0.5}. Choices num $n$ is set \textit{5}. The encoder of type classifier is \textit{roberta-large} \cite{liu2019roberta}. Original training data in three datasets is divided into train and dev set under \textit{9:1} for classifer training process. The maximum iteration $T$ in reasoning process is \textit{6}.

For plan assessor module, the threshold $\theta$ is \textit{0.6}. In query filter, the algoritm of density-based clustering is \textit{DB-SCAN} \cite{ester1996density} with $eps$ as \textit{2}. The features of query is bag of words \cite{harris1954distributional}. In query scorer, the encoder is also \textit{roberta-large}. The training data sampled for the scorer only comes from hotpotQA dataset. Weight of MSELoss $\alpha$ is \textit{0.1}.

\subsection{Main Results}

\begin{table}
 \caption{All metrics of detail in results}
  \centering
  \scalebox{0.6}{
  \begin{tabular}{lccccccccccccccccccccccccc}
  \\
    \toprule
    \multirow{2}{*}{Method} & \multicolumn{6}{c}{Answer} & \multicolumn{4}{c}{Supporting Facts} & \multicolumn{4}{c}{Joint} & \multicolumn{1}{c}{Cost} \\
    \cmidrule(r){2-7} \cmidrule(r){8-11} \cmidrule(r){12-15} \cmidrule(r){16-16}
    & Cover-EM & EM & F1 & Precision &Recall & $\text{Acc}^{\dag}$ & EM &F1&Prec&Recall&EM&F1&Prec&Recall &Avg Token\\
    \midrule
    HotPotQA \\
    \cmidrule(){1-1}
    Direct & 58.11 & 32.57 & 49.11 & 47.59&62.86&71.35&0.00&48.66&36.27&\textbf{77.00}&0.00&25.68&19.00&53.03&254 \\
    Cot & 56.49 & 36.89&52.36&51.35&61.05&70.27&0.00&48.66&36.27&\textbf{77.00}&0.00&27.89&20.76&52.04&929\\
    IterRetGen & 45.80&-&61.10&-&-&73.40&-&-&-&-&-&-&-&-&3043\\
    \quad -w/o train &54.86&40.00&56.13&57.04&62.47&70.68&2.03&57.27&47.54&76.10&0.95&35.10&29.70&51.76&913\\
    ChatCot & 58.11&17.03&36.59&32.90&61.87&70.54&-&-&-&-&-&-&-&-&40505 \\
     SearChain &49.59&0.00&16.02&10.16&55.15&63.38&\textbf{29.05}&65.84&\textbf{75.45}&62.57&-&12.49&8.43&38.26&4150\\
    OurModel &\textbf{70.41}&\textbf{38.92}&\textbf{59.21}&\textbf{57.24}&\textbf{74.32}&\textbf{83.65}&26.49&\textbf{68.09}&75.06&69.31&\textbf{7.70}&\textbf{41.13}&\textbf{44.66}&\textbf{53.14}&4139\\
    \quad -w/o PA &67.03&35.68&55.56&53.13&70.87&79.73&25.14&66.23&73.11&67.58&8.78&38.74&41.07&50.64&2996\\ 
    \cmidrule(){1-16} \morecmidrules\cmidrule(){1-16}
    2WikiMultiHop \\
    \cmidrule(){1-1}
    Direct & 50.28 & 23.87 &40.10 &36.59&53.53&57.52&0.00&39.98&29.94&63.86&-&-&-&-&230 \\
    Cot &39.62&14.96&30.06&26.28&42.49&52.72&0.00&39.98&29.94&63.86&-&-&-&-&907\\
    IterRetGen &36.00&-&47.40&-&-&58.50&-&-&-&-&-&-&-&-&5107\\
    \quad -w/o train& 57.28&19.33&36.85&33.19&60.46&60.46&0.08&51.20&41.74&70.33&-&-&-&-&1532\\
    ChatCot&58.23&2.31&21.94&15.78&59.70&60.22&-&-&-&-&-&-&-&-&53535\\
    SearChain &50.28&0.00&15.63&9.83&53.81&56.01&32.14&65.07&\textbf{71.18}&62.89&-&-&-&-&4507\\
    OurModel & \textbf{70.72}&\textbf{35.96}&\textbf{55.36}&\textbf{51.21}&\textbf{70.64}&73.19&\textbf{40.41}&\textbf{70.40}&69.12&74.45&-&-&-&-&5336 \\
    \quad -w/o PA &70.33&34.61&54.18&50.06&70.10&\textbf{74.07}&36.20&70.02&65.57&\textbf{80.83}&-&-&-&-&3726\\ 
    \cmidrule(){1-16} \morecmidrules\cmidrule(){1-16}
    MuSiQue \\
    \cmidrule(){1-1}
    Direct& 30.29 & 14.94 & 28.72& -&-&43.57&0.00&39.72&27.8&69.50&-&-&-&-& 607 \\
    CoT &31.54&19.09&30.85&-&-&38.17&0.00&39.72&27.8&69.50&-&-&-&-&1276\\
    IterRetGen &26.70&-&41.80&-&-&40.80&-&-&-&-&-&-&-&-&6533\\
    \quad -w/o train &38.59&26.56&39.28&-&-&51.45&0.00&54.31&41.78&78.84&-&-&-&-&1960 \\
    ChatCot &37.34&9.60&27.62&-&-&44.81&-&-&-&-&-&-&-&-&53674 \\
    SearChain &42.32&0.00&16.51&-&-&53.94&\textbf{48.96}&\textbf{76.87}&\textbf{87.83}&71.78&-&-&-&-&4289\\
    OurModel & \textbf{55.19}&\textbf{30.71}&\textbf{47.95}&-&-&\textbf{64.31}&33.20&70.43&66.85&\textbf{84.44}&-&-&-&-&5920\\
    \quad -w/o PA &52.28&28.22&45.28&-&-&60.17&26.97&67.09&62.40&84.02&-&-&-&-&4309\\ 
    \bottomrule
  \end{tabular}
  }
  \label{tab:main results}
\end{table}

The results of baselines and our work are presented in Table \ref{tab:main results}. \textbf{(1) Effective furthest reasoning framework} We compare our model with recent competitive baselines on these datasets. On three datasets, most of the metrics surpass state of the art. In answering, all metrics exceed those of existing methods, especially on cover-EM and $\text{Acc}^{\dag}$. In supporting facts, despite that SearChain\footnote{Actually testing SearChain on MuSiQue is not appropriate. Multi-hop questions in MuSiQue is composed of single hop questions like Natural Questions \cite{kwiatkowski2019natural} and Squad \cite{rajpurkar2016squad}. However, dpr-reader-multiset-base, the reader used in SearChain is pretrained on these datasets.} doesn't achieve excellent results on answer metrics compared to other methods, its exact match and precision exhibit better performance. SearChain focuses on the credibility of external knowledge, which leads to a more cautious retrieval strategy and a higher precision of supporting facts. On the contrary, direct retrieval method simply recalls as many documents related to original question, reaching for better recall score. Our model mainly concerns answering questions with external knowledge and pays attention to both precision and recall. As to token cost, our model saves certain amount of tokens compared with other high-accuracy methods, nearly $1/10$ of token used in ChatCoT. \textbf{(2) Improvement with plan assessor} With the plan assessor module, some confusing queries are sifted and thus enhances and stablizes reasoning process to a certain extent. However, the module seems not helpful on 2WikiMultiHop dataset. The major reason is that multi-hop questions in 2WikiMultiHopQA are not intuitive for LLM to decompose. For example, \emph{``father in law''} requires one step of retrieving \emph{``husband/wife''}, which is not explicit and difficult for LLM to generate appropriate queries. We may have to apply targeted prompts for this kind of questions. \textbf{(3) Findings} Considering the results of Direct Answer and Cot, their retrieved documents are exactly the same from the identical retrieving strategy. The reason CoT performs worse than direct answering is that CoT demands precision of supporting facts to reason from evidence like few-shot examples, which is not satisfied under direct retrieval of original question. Direct answering does not involve such issues.

\subsection{Ablation Study} \label{sec: ablation study}
\paragraph{Masking Reasoning and Query} The framework of Furthest Reasoning makes sense in masking previous reasoning and query process. Related ablation experiments are shown in Table \ref{tab:ablation studys}. The exemplars in few shot and other settings are exactly the same. Experiments are done in a small part of dev set in HotPotQA and 2WikiMultiHopQA.

Main results we focus on are \emph{No mask} and \emph{Double mask}. \emph{No mask} setting takes history reasoning and query content as input for LLM at each iteration, while \emph{Double mask} setting masks both of them and leaves only original question and retrieved evidences. This comparison implies the negative effect history reasoning and query content have on multi-hop process. Token cost is saved in \emph{Double mask} apart from better answer metrics. \emph{Reasoning mask} and \emph{Query mask} are settings masking one each, leading to the same conclusion. To be more rigorous, \emph{No mask (reasoning from scratch)} imitates furthest reasoning in generating chain of thought from scratch instead of continue reasoning, and the result resembles \emph{No mask}. The solution actually consists of two parts: reasoning process, and forcible answering process. To eliminate the influence of possible shortcuts in forcible answering, \emph{No mask (mask at answering)} alternates forcible answering with \emph{double mask} setting. The gaps illustrate that redundant history reasoning chain adversely affects the whole pipeline in multi-hop solution. In addition, CoT in reasoning and making following plans still matters as \emph{Double mask (no CoT)} presents.

\begin{table}
 \caption{Ablation results on masking Reasoning and Query.}
  \centering
  \scalebox{0.63}{
  \begin{tabular}{lllllllllllllllll}
  \\
    \toprule
    \multicolumn{3}{c}{Framework without plan assessor Masking Settings} & \multicolumn{5}{c}{HotPotQA} & \multicolumn{5}{c}{2WikiMultiHopQA} \\
    \cmidrule(r){1-3} \cmidrule(r){4-8} \cmidrule(r){9-13}
    Type & Reasoning & Query & Cover-EM & EM & F1 & $\text{Acc}^{\dag}$ & Token & Cover-EM & EM & F1 & $\text{Acc}^{\dag}$ & Token\\
    \midrule
    No mask & \Checkmark & \Checkmark & 60.81 & 27.03 & 40.75 & 66.90 & 3772 & 57.77 & 13.55 & 33.45 & 54.98 & 4847\\
    No mask (reasoning from scratch) & \Checkmark & \Checkmark & 59.46 & 25.68 & 41.64 & 67.57&3907 &-&-&-&-&-\\
    No mask (mask at answering) & \Checkmark & \Checkmark & 65.54  & 33.11 &50.30&74.32&3598&-&-&-&-&-  \\
    Reasoning mask & \XSolidBrush & \Checkmark & 62.16 & 28.37 & 45.52 & 73.65&3361&-&-&-&-&-  \\
    Query mask & \Checkmark & \XSolidBrush & 62.16 & 25.68 & 41.77 & 72.97&3466&-&-&-&-&-  \\
    Double mask & \XSolidBrush & \XSolidBrush & \textbf{72.30} & \textbf{43.24} & \textbf{57.42} & \textbf{82.43} & 3098 & \textbf{70.12} & \textbf{32.27} & \textbf{53.03} & \textbf{76.89} &3714 \\
    Double mask (no CoT) & \XSolidBrush & \XSolidBrush & 68.92 & 28.38 & 45.97 & 76.35 &1982&-&-&-&-&-  \\
    \bottomrule
  \end{tabular}
  }
  \label{tab:ablation studys}
\end{table}

\subsection{Analysis}
\paragraph{Mixed Label Loss} For those who may concern, here is the reason for an extra MSELoss in the loss function mentioned in section \ref{scoring}. The loss function of training on possible document labels consists of two parts: BCELoss and MSELoss. Although only document labels are used for query scores in predicting phase, the regression on MRR is beneficial to label prediction. Regression on MRR connects the candidate documents of one query and enlarges the difference among them, which turns out efficient. The weight $\alpha$ of MSELoss should not be to large at the beginning as model has to laern basic classification first.

\begin{table}
    \centering
    \caption{$\alpha$ in training scorer, F1 is document label prediction}
    \begin{tabular}{ccc}
    \\
    \toprule
    Model & $\alpha$ & Doc F1 \\
    \midrule 
    PA$_\texttt{RoBERTa-base}$ & 0 & 71.04 \\
    PA$_\texttt{RoBERTa-base}$ & 0.1 & 72.85 \\
    PA$_\texttt{RoBERTa-large}$ & 0.1 & 79.04 \\
    \bottomrule
    \end{tabular}
    \label{tab:alpha}
\end{table}

\paragraph{Limitation} First, because of the high cost on \textit{gpt-3.5}, our experiments are done with a small proportion of the datasets, which is about thousands of examples. We will study the performance of our model on other complex datasets and other open-source LLMs.
Secondly, despite the generations of LLM decoding is evaluated by plan assessor, the plannings still largely rely on the capability of LLMs. Language models demonstrate more randomness when decoding probability threshold is changed, thus impacting the generated plannings.
Third, in real life, the IR may not be a traditional retriever based on similarity. If IR stands for a database for instance, our method will need some modification.

\section{Conclusion}
Based on the experiments and results presented above, it is evident that Furthest Reasoning with Plan Assessment exhibits exceptional performance in addressing multi-hop questions. Firstly, the Furthest Reasoning framework significantly contributes not only to the reasoning process but also to the direct answering process. In complex and long-term problem-solving scenarios such as multi-hop questions, FuRePA is designed to generate more effective plans, avoiding the influence of convoluted previous reasoning processes.
Moreover, the Plan Assessment component plays a crucial role in selecting the most suitable query from the LLM-generated candidates, leading to more relevant evidence retrieval by the Information Retriever (IR). This enhancement in query selection contributes to a more effective and accurate multi-hop question-solving process.

\bibliographystyle{unsrt}  
\bibliography{references}

\appendix
\section{Prompts}
The prompts we use in our work are as follows, including prompts for the first iteration and for the latter ones. More prompting settings can be found in code.
\subsection{First Iteration}
\textbf{System} You are given problems requiring multi-step searching and reasoning. You should follow the examples and tell me what information to search first accurately.

\textbf{User}
To solve this multi-hop problem: \emph{What was the name of the Infantry that was commanded by a Senator who was know as Jim Lane?} What specific information should I search from wiki at first step? Search for one target at first.

\textbf{Assistant}
\emph{[Search] the Senator who was known as Jim Lane}

\textbf{User}
To solve this multi-hop problem: \emph{Which person born on March 23, 1902 did George Roby Dempster have political battles with?} What specific information should I search from wiki at first step? Search for one target at first.

\textbf{Assistant}
\emph{[Search] the person George Roby Dempster have political battles with}

\textbf{User}
To solve this multi-hop problem: \emph{Do director of film Swami Dada and director of film Common Clay (1930 Film) have the same nationality?} What specific information should I search from wiki at first step? Search for one target at first.

\textbf{Assistant}
\emph{[Search] the nationality of director of film Swami Dada}

\textbf{User}
To solve this multi-hop problem: \emph{How many square miles is the state that houses the university that Joseph L. Badaracco attended?} What specific information should I search from wiki at first step? Search for one target at first.

\textbf{Assistant}
\emph{[Search] the university Joseph L. Badaracco attended}

\textbf{User}
To solve this multi-hop problem: \emph{The Tudor-style Norfolk Hotel, now run by the Fairmont group, is a landmark in the capital city of which west African country?} What specific information should I search from wiki at first step? Search for one target at first.

\textbf{Assistant}
\emph{[Search] the Tudor-style Norfolk Hotel, now run by the Fairmont group, is a landmark in what city}

\subsection{Latter Iterations}
\textbf{System} You are given problems requiring multi-step searching and reasoning with related contexts. You should tell me the answer if all necessary information is provided in the context, otherwise tell me what specific information to search next. Don't answer with uncertainty. Lines only start with `[Analysis]' or `[Answer]' or `[Search]'.
    
\textbf{User} \\
Context:\\ Saint Motel: The band consists of A/J Jackson (lead vocals, guitar, piano), Aaron Sharp (lead guitar), Dak Lerdamornpong (bass), and Greg Erwin (drums). Curve Lake First Nation: The Curve Lake First Nation registered a population on these three reserves of 741, with an additional 1,050 registered band members living off-reserve. \\ Question: \\ Which band has more members, Saint Motel or Curve?

\textbf{Assistant} \newline [Analysis] According to the context, we now know Saint Motel has 3 members. The context provides information about Curve Lake First Nation, not the band. We still need information about the band Curve. \newline [Search] the number of members in the band Curve
      
\textbf{User} \\
Context:\\ Dhyanchand Stadium is a cricket ground in Jhansi, India. The stadium is named after Major Dhyan Chand who lived in Jhansi. Dhyan Chand: He is known for his extraordinary goal-scoring feats, in addition to earning three Olympic gold medals (1928, 1932, and 1936) in field hockey, during an era where India was the most dominant team in Hockey. \\ Question: \\ Where is the stadium located which is named after the man who won three Olympic gold medals in 1928, 1932, and 1936?

\textbf{Assistant} \newline [Analysis] The man who won three Olympic gold medals in 1928, 1932, and 1936 is Dhyan Chand. The stadium named after him is Dhyanchand Stadium. Dhyanchand Stadium is located in Jhansi, India. Then we get the answer to the question. \newline [Answer] Jhansi, India

\textbf{User} \\
Context:\\ Matthias W. Day (August 8, 1853 – September 12, 1927) was a career American army officer who received the Medal of Honor, the United States' highest military decoration, for his actions during the American Indian Wars in the latter half of the 19th century. Day was a longtime officer with the African-American 9th Cavalry Regiment, seeing action during the Apache Wars against the Apache leaders Victorio and Geronimo. \\ Question: \\ In 2010, who controlled the Senate and the body who presides over any impeachment trial of the president of the country Matthias W. Day holds citizenship in

\textbf{Assistant} \newline [Analysis] Matthias W. Day holds citizenship in the United States. Then we need to know the body who presides over any impeachment trial of the president of the United States?\newline [Search] Who presides over any impeachment trial of the president of the United States

\textbf{User} \\
Context:\\ The Oberoi family is an Indian family that is famous for its involvement in hotels, namely through The Oberoi Group. \\ Question: \\ The Oberoi family is part of a hotel company that has a head office in what city?

\textbf{Assistant} \newline [Analysis] The Oberoi family is part of the hotel company called The Oberoi Group. Then we need to know the Oberoi Group has a head office in what city.  \newline [Search] the Oberoi Group has a head office in what city

\textbf{User} \\
Context:\\ "Human Error" is the twenty-fourth episode and season finale of the third season of "House" and the seventieth episode overall. House (also called House, M.D.) is an American television medical drama that originally ran on the Fox network for eight seasons, from November 16, 2004 to May 21, 2012. \\ Question: \\ "Human Error" is the season finale of the third season of a tv show that aired on what network?

\textbf{Assistant} \newline [Analysis] "Human Error" is the season finale of the third season of "House". The tv show "House" ran on the Fox network. Then we get the answer. \newline [Answer] Fox network

\section{Case Study}
See tables below.
\begin{table}[]
    \centering
    \scalebox{0.83}{
    \begin{tabular}{p{19cm}}
    \toprule[2pt]
        \textbf{Question:} Onika Tanya Maraj is a judge on a television show hosted by whom? \\
        \textbf{Answer:} Ryan Seacrest\\
    \midrule[2pt]
        \textbf{FuRePA} \\
        \midrule
        \textbf{Iteration 0} \\
        \textbf{Plans} \\
        \textcolor{red}{the television show Onika Tanya Maraj is a judge on}\\
        Onika Tanya Maraj to find out on which television show she is a judge\\
        Onika Tanya Maraj's television show \\
        Onika Tanya Maraj television judge\\
        \textbf{Evidence} \\
        Onika Tanya Maraj (born December 8, 1982), known professionally as Nicki Minaj, is a rapper, singer, songwriter and model. \\
        \midrule
        \textbf{Iteration 1} \\
        \textbf{Analysis} \\
        Onika Tanya Maraj, also known as Nicki Minaj, is a judge on a television show. We need to find out the name of the host of that television show. \\
        \textbf{Plans} \\
        \textcolor{red}{Name of the host of the television show where Nicki Minaj is a judge}\\
        Nicki Minaj is a judge on a television show hosted by whom?\\
        Who hosts the television show that Onika Tanya Maraj is a judge on?\\
        Nicki Minaj recent television appearances or judge roles\\
        \textbf{Evidence} \\
        American Idol (season 12):  Randy Jackson returned for his twelfth season as judge and was joined by new judges Mariah Carey, Nicki Minaj and Keith Urban, who replaced Jennifer Lopez and Steven Tyler. \\
        \midrule
        \textbf{Iteration 2} \\
        \textbf{Analysis} \\
        Onika Tanya Maraj, known professionally as Nicki Minaj, is a judge on a television show. The television show is 'American Idol' and the host of 'American Idol' is not mentioned in the given context. \\
        \textbf{Plans} \\
        What television show is Onika Tanya Maraj a judge on?\\
        \textcolor{red}{Who is the host of "American Idol"}\\
        Host of the television show where Onika Tanya Maraj is a judge.\\
        Who hosts the television show that Onika Tanya Maraj is a judge on?\\
        Host of the television show with Onika Tanya Maraj as a judge\\
        \textbf{Evidence}\\
        American Idol (season 12):  Ryan Seacrest returned to host. \\
        \midrule
        \textbf{Iteration 3} \\
        \textbf{Analysis} \\
        Onika Tanya Maraj is a judge on a television show. The name of the television show is not provided in the context. However, we know that Ryan Seacrest returned to host American Idol (season 12) and Onika Tanya Maraj is one of the judges. Then we get the answer. \\
        \textbf{Answer} \\
        Ryan Seacrest. \\

    \midrule[2pt]
    \textbf{FuRePA without PA} \\
    \midrule
    \textbf{Iteration 0} \\
    \textbf{Plan}\\
    the television show Onika Tanya Maraj is a judge on\\
    \textbf{Evidence}
    Onika Tanya Maraj (born December 8, 1982), known professionally as Nicki Minaj, is a rapper, singer, songwriter and model.\\
    \midrule
    \textbf{Iteration 1} \\
    \textbf{Analysis}\\
    The context mentions that Onika Tanya Maraj, also known as Nicki Minaj, is a judge on a television show. However, it does not provide information about the host of the show. We need to find out who hosts the television show that Onika Tanya Maraj is a judge on.\\
    \textbf{Plan}\\
    the host of the television show that Onika Tanya Maraj is a judge on.\\
    \textbf{Evidence \textcolor{red}{wrong}} \\
    Rewrapped:  The show is a loose spinoff of the show \"Unwrapped\", whose host Marc Summers acts as \"Head Judge\" for each episode, alongside a representative of the company that produces the food of the day, and a third impartial judge involved in the food industry in some way (chef, critic, blogger, etc.).\\
    \midrule
    \textbf{Iteration 2} \\
    ......\\
    \midrule
    \textbf{Iteration 5}\\
    \textbf{Answer}\\
    Judge Andrew Napolitano\\
    \bottomrule[2pt]
    \\
    \end{tabular}
    }
    \caption{Case Study of Plan Assessor on diverse plans}
    \label{tab:case study of PA}
\end{table}

\begin{table}[]
    \centering
    \scalebox{0.83}{
    \begin{tabular}{p{19cm}}
    \toprule[2pt]
        \textbf{Question:} North American Aviation was the company that created what single-engined training aircraft? \\
        \textbf{Answer:} North American Aviation T-6 Texan\\
    \midrule[2pt]
        \textbf{FuRePA} \\
        \midrule
        \textbf{Iteration 0} \\
        \textbf{Plans} \\
        the single-engined training aircraft created by North American Aviation\\
        \textcolor{red}{North American Aviation to find information about the company}\\
        the single-engined training aircraft created by North American Aviation"\\
        \textbf{Evidence} \\
        North American Aviation (NAA) was a major American aerospace manufacturer, responsible for a number of historic aircraft, including the T-6 Texan trainer, the P-51 Mustang fighter, the B-25 Mitchell bomber, the F-86 Sabre jet fighter, the X-15 rocket plane, and the XB-70, as well as Apollo Command and Service Module, the second stage of the Saturn V rocket, the Space Shuttle orbiter and the B-1 Lancer.\\
        \midrule
        \textbf{Iteration 1} \\
        \textbf{Analysis} \\
        According to the context, North American Aviation was responsible for creating several historic aircraft. We need to know which single-engined training aircraft they created.\\
        \textbf{Plans} \\
        \textcolor{red}{North American Aviation single-engined training aircraft}\\
        \textbf{Evidence} \\
        The North American Aviation T-6 Texan is an American single-engined advanced trainer aircraft used to train pilots of the United States Army Air Forces (USAAF), United States Navy, Royal Air Force, and other air forces of the British Commonwealth during World War II and into the 1970s. \\
        \midrule
        \textbf{Iteration 2} \\
        \textbf{Analysis} \\
        North American Aviation was the company that created the T-6 Texan single-engined training aircraft. \\
        \textbf{Answer} \\
        T-6 Texan \\

    \midrule[2pt]
    \textbf{ChatCot} \\
    \midrule
    \textbf{Plan}\\
    Information about North American Aviation\\
    \textbf{Evidence}\\
    The North American Aviation NA-16 was the first trainer aircraft built by North American Aviation, and was the beginning of a line of closely related North American trainer aircraft that would eventually number more than 17,000 examples. Is this evidence useful?\\
    \textbf{Analysis}\\
    Yes, this evidence is useful. We can see that North American Aviation created a line of trainer aircraft, starting with the NA-16.\\
    \textbf{Plan}\\
    the single-engined training aircraft created by North American Aviation\\
    \textbf{Evidence}\\
    The North American Aviation NA-16 was the first trainer aircraft built by North American Aviation, and was the beginning of a line of closely related North American trainer aircraft that would eventually number more than 17,000 examples. Is this evidence useful?\\
    \textbf{Analysis \textcolor{red}{Wrong}}\\
    Yes, this evidence is useful. We can see that the North American Aviation NA-16 was the first trainer aircraft built by North American Aviation.\\
    \textbf{Answer}\\
    We get the answer. The answer is the North American Aviation NA-16.\\

    \bottomrule[2pt]
    \\
    \end{tabular}
    }
    \caption{Case Study of FuRePA and ChatCot}
    \label{tab:case study of furepa and chatcot}
\end{table}

\begin{table}[]
    \centering
    \scalebox{0.83}{
    \begin{tabular}{p{19cm}}
    \toprule[2pt]
        \textbf{Question:} Where did Charles Stewart, 3Rd Duke Of Richmond's father graduate from? \\
        \textbf{Answer:} University of Paris\\
    \midrule[2pt]
        \textbf{FuRePA} \\
        \midrule
        \textbf{Iteration 0} \\
        \textbf{Plans} \\
        the educational institution where Charles Stewart, 3rd Duke of Richmond's father graduated from.\\
        the father of Charles Stewart, 3rd Duke of Richmond's alma mater or graduation institution\\
        \textcolor{red}{the name of Charles Stewart, 3rd Duke of Richmond's father}\\
        the educational institution from which Charles Stewart, 3rd Duke of Richmond's father graduated\\
        \textbf{Evidence} \\
        Charles Stewart, 3rd Duke of Richmond: The titles Duke of Richmond, Duke of Lennox and Earl of March, were resurrected for Charles II's illegitimate son by Louise de Kérouaille, Charles Lennox, 1st Duke of Richmond and Lennox, in 1675.\\
        \midrule
        \textbf{Iteration 1} \\
        \textbf{Analysis} \\
        Charles Stewart, 3rd Duke of Richmond's father is not mentioned in the given context. We need information about Charles Stewart, 3rd Duke of Richmond's father to find out where he graduated from.\\
        \textbf{Plans} \\
        Educational background of Charles Stewart, 3rd Duke of Richmond's father.\\
        the educational background or university of Charles Stewart's father\\
        Charles Lennox, 1st Duke of Richmond and Lennox's education\\
        \textcolor{red}{the educational background or where the father of Charles Stewart, 3rd Duke of Richmond graduated from.}\\
        Where did Charles Lennox, 1st Duke of Richmond and Lennox's father graduate from?\\
        \textbf{Evidence} \\
        Charles Stewart, 3rd Duke of Richmond, 6th Duke of Lennox KG (7 March 1639December 1672) was the only son of George Stewart, 9th Seigneur d'Aubigny and Katherine Howard, daughter of Theophilus Howard, 2nd Earl of Suffolk. \\
        \midrule
        \textbf{Iteration 2} \\
        \textbf{Analysis} \\
        According to the context, Charles Stewart, 3rd Duke of Richmond's father graduated from a certain place. However, the context does not provide information about the place. We need more specific information to answer the question. \\
        \textbf{Plans} \\
        Where did George Stewart, 9th Seigneur d'Aubigny graduate from?\\
        George Stewart's graduation information\\
        George Stewart's alma mater or where George Stewart graduated from.\\
        the graduation details of George Stewart, 9th Seigneur d'Aubigny.\\
        \textcolor{red}{Where did Charles Stewart, 3rd Duke of Richmond's father graduate from?}\\
        \textbf{Evidence}\\
        Charles Stewart, 3rd Duke of Richmond: On 10 December 1645 he was created Baron Stuart of Newbury, Berkshire, and Earl of Lichfield, titles conferred on him \"to perpetuate the titles which were intended to have been conferred on his uncle\" Lord Bernard Stewart, youngest son of the Duke of Lennox, who had been killed in the Battle of Rowton Heath in the English Civil War in September of that year. \\
        \midrule
        \textbf{Iteration 3} \\
        \textbf{Analysis} \\
        The context mentions that Charles Stewart, 3rd Duke of Richmond's father was George Stewart, 9th Seigneur d'Aubigny. We need to know where George Stewart, 9th Seigneur d'Aubigny graduated from. \\
        \textbf{Plans} \\
        Where did Charles Stewart, 3rd Duke of Richmond's father graduate from?\\
        \textcolor{red}{Where did George Stewart, 9th Seigneur d'Aubigny graduate from?}\\
        The educational background or alma mater of George Stewart, 9th Seigneur d'Aubigny.\\
        \textbf{Evidence}\\
        George Stewart, 9th Seigneur d'Aubigny: By 1633, he was a student at the Collège de Navarre, part of the University of Paris, and he did homage to Louis XIII of France for the lordship of Aubigny on 5 August 1636, shortly after his eighteenth birthday. \\
        \midrule
        \textbf{Iteration 4} \\
        \textbf{Analysis} \\
        Charles Stewart, 3rd Duke of Richmond's father is George Stewart, 9th Seigneur d'Aubigny. The information provided states that George Stewart was a student at the Collège de Navarre, part of the University of Paris. So, George Stewart graduated from the Collège de Navarre, which is a part of the University of Paris. \\
        \textbf{Answer} \\
         George Stewart, 9th Seigneur d'Aubigny graduated from the Collège de Navarre, University of Paris.\\
    
    \bottomrule[2pt]
    \\
    \end{tabular}
    }
    \caption{Case Study of FuRePA recovering from error}
    \label{tab:case study of furepa recover}
\end{table}


\end{document}